\documentclass{article}

\usepackage[final]{corl_2020} 

\usepackage{hyperref}
\usepackage{xcolor}
\usepackage{amsmath}
\usepackage{amssymb}
\usepackage{mathtools}
\usepackage{siunitx}
\usepackage{graphicx}
\usepackage{adjustbox}
\usepackage{float}
\usepackage{wrapfig}
\usepackage[labelformat=simple,font=footnotesize]{subcaption}
\usepackage[font=small,skip=3pt,tableposition=top]{caption}
\usepackage{booktabs}
\usepackage{bbm}
\usepackage{multirow}
\usepackage[ruled,vlined]{algorithm2e}

\title{Learning RGB-D Feature Embeddings for Unseen Object Instance Segmentation}

\author{ Yu Xiang$^1$ \hspace{4px} Christopher Xie$^2$ \hspace{4px}  Arsalan Mousavian$^1$ \hspace{4px} Dieter Fox$^{1, 2}$\\
$^1$NVIDIA \hspace{6px} $^2$University of Washington\\
\tt\small \{yux,amousavian,dieterf\}@nvidia.com \hspace{2px} chrisxie@cs.washington.edu 
}

\begin{document}
\maketitle


\begin{abstract}
Segmenting unseen objects in cluttered scenes is an important skill that robots need to acquire in order to perform tasks in new environments. In this work, we propose a new method for unseen object instance segmentation by learning RGB-D feature embeddings from synthetic data. A metric learning loss function is utilized to learn to produce pixel-wise feature embeddings such that pixels from the same object are close to each other and pixels from different objects are separated in the embedding space. With the learned feature embeddings, a mean shift clustering algorithm can be applied to discover and segment unseen objects. We further improve the segmentation accuracy with a new two-stage clustering algorithm. Our method demonstrates that non-photorealistic synthetic RGB and depth images can be used to learn feature embeddings that transfer well to real-world images for unseen object instance segmentation.
\end{abstract}



\section{Introduction}
\label{sec:intro}

In order to perform complex tasks in different environments autonomously, robots need to learn various skills in perception, planning and control. Among the perception skills, the ability to segment unseen objects in cluttered scenes is critical. Imagine that a robot is tasked to clean a kitchen. It may encounter objects it has never seen before. Object recognition approaches that rely on building 3D models of objects \cite{xiang2017posecnn,tremblay2019deep} cannot be applied to these scenarios.

To segment unseen objects, robots need to learn the concept of an ``object'' and generalize it to new objects. This is usually achieved by training an object perception method with many different objects. However, there is no large scale dataset of real images that contains many objects in robotic manipulation scenes. Existing large scale datasets such as ImageNet \cite{deng2009imagenet} or COCO \cite{lin2014microsoft} are collected from web images, which are quite different from robotic manipulation settings such as tabletop scenes. In addition, there is no depth image available in these datasets, while depth information has proven to be very useful in robotic manipulation \cite{mahler2017dex,mousavian20196}.

As a result, previous works on Unseen Object Instance Segmentation (UOIS) \cite{xie2019the,danielczuk2019segmenting} utilize synthetic data for training. 3D CAD models of objects are used to compose a scene that is rendered into RGB-D images from different viewpoints. The benefit of using synthetic data is that a large number of images can be generated. For example, \cite{xie2019the} generated 40,000 scenes with 7 RGB-D images per scene. However, models trained on synthetic data may not work well on real images due to the sim-to-real domain gap, especially when the RGB images are non-photorealistic. A common recipe to solve this sim-to-real problem is to use depth images for training as shown in \cite{xie2019the,danielczuk2019segmenting}. Adding non-photorealistic RGB images in training typically degrades performance for these methods.

In this work, we propose a new method for UOIS by learning RGB-D feature embeddings directly from synthetic data. Different from previous methods, we show that our method is able to utilize both depth images and non-photorealistic RGB images to produce feature embeddings for every pixel, which can be used in a clustering algorithm to segment unseen objects. More importantly, adding the non-photorealistic RGB images in our training improves the segmentation accuracy due to our metric learning formulation.

Specifically, a fully convolutional network is used to process an RGB-D image to produce a dense feature map with the same size as the input image. In training, our goal is to learn to produce feature embeddings such that pixels from the same object are close to each other, and pixels from different objects are separated in the embedding space. To achieve this goal, we apply a metric learning loss based on cosine distance to the unit-length normalized feature embeddings. In testing, we utilize the von Mises-Fisher mean shift algorithm~\cite{kobayashi2010mises} to cluster these feature embeddings, which automatically discovers the number of objects in the image and generates a segmentation mask for each object. In addition, we introduce a two-stage clustering process to further improve the segmentation accuracy. The first stage clusters all the pixels from the input image. Then, for each segmentation mask from the first stage, a Region of Interest (RoI) of the mask is extracted from the input image and the second stage clustering operation is applied to it. The second stage clustering can refine the segmentation boundaries and separate objects that are close to each other if the first stage clustering fails to separate them. We conduct experiments on two real-world RGB-D image datasets \cite{richtsfeld2012segmentation,suchi2019easylabel} to evaluate our method named Unseen Clustering Network (UCN), and demonstrate the state-of-the-art results on these datasets.

In summary, our contributions in this work are i) we show that learning RGB-D feature embeddings with a metric learning formulation can better utilize the synthetic RGB images even if they are non-photorealistic; ii) we experimentally investigate different ways of fusing RGB and Depth; iii) we introduce a new two-stage clustering algorithm for UOIS; iv) our method achieves the state-of-the-art performance on two commonly used datasets for UOIS.

\vspace{-2mm}
\section{Related Work}
\label{sec:related}

\textbf{Unseen Object Instance Segmentation.} Earlier works for UOIS apply image segmentation methods to segment objects such as GraphCut~\cite{felzenszwalb2004efficient}, Connected Components~\cite{trevor2013efficient}, LCCP~\cite{christoph2014object} and SceneCut~\cite{pham2018scenecut}. These methods are based on low-level image cues such as edges, contours, connectivity, convexity or symmetry to group pixels into objects. Consequently, they tend to segment every thing in an image and objects are usually over-segmented due to textures or concave shapes. This is mainly because there is no object-level supervision, and these methods cannot learn the concept of object.

On the other hand, recent learning-based approaches generate synthetic data to train neural networks for unseen object segmentation \cite{shao2018clusternet,danielczuk2019segmenting,xie2019the}, where a large number of images can be rendered using 3D CAD models. By providing object-level supervision, these methods achieve better performance than image-segmentation-based methods. However, using synthetic data for training requires dealing with the sim-to-real gap. Previous works~\cite{danielczuk2019segmenting,xie2019the} rely on using depth images to bridge the sim-to-real gap. We show that our method can also utilize the non-photorealistic RGB images to learn a feature representation for effectively segmenting unseen objects.

\textbf{Deep Metric Learning.} Recently, combining deep neural networks and metric learning to learn powerful feature representations has been successful in many applications such as image recognition~\cite{chen2020simple}, image clustering~\cite{oh2016deep}, face recognition~\cite{schroff2015facenet}, reinforcement learning~\cite{srinivas2020curl} and robotic manipulation~\cite{florence2018dense}. Most works focus on learning visual representations using real images. In this work, we show that it is possible to learn feature embedding networks directly from synthetic RGB-D data that robustly transfer to real images for segmenting unseen objects, even if the synthetic RGB images are non-photorealistic.

\vspace{-2mm}
\section{Method}

Given an RGB-D image, our goal is to segment individual objects in the image, where these objects are unseen during training. Since the concept of ``objects`` is quite general, in this work, we consider objects in robotic manipulation environments and focus on segmenting objects in tabletop scenes, which provides useful information for manipulating these objects as shown in~\cite{mousavian20196,murali20196}.

\subsection{Learning from Synthetic Data}

\begin{figure*}
	\centering
	\includegraphics[height=0.25\textwidth,width=\textwidth]{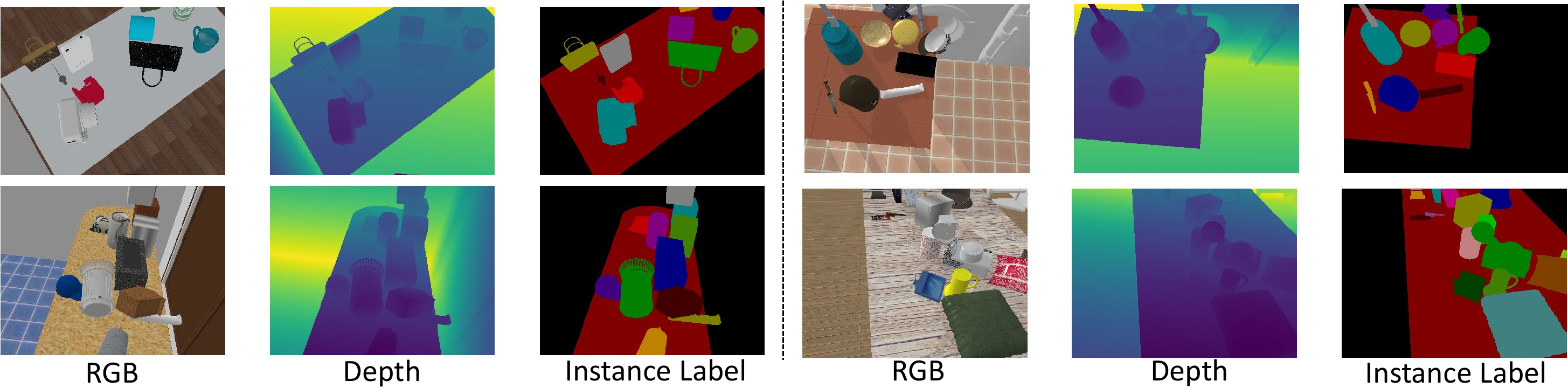}
	\caption{Example RGB-D images and the corresponding instance labels from the Tabletop Object Dataset~\cite{xie2019the}.}
	\label{fig:tabletop}
\end{figure*}

We employ deep neural networks for unseen object instance segmentation. In order to segment unseen objects, a network needs to learn the concept of objects and be able to generalize it to new objects. We achieve this by training the network with a large number of objects in tabletop scenes. However, there does not exist a large scale dataset of real-world images for many objects in tabletop scenes. Therefore, we resort to using synthetic data for training, where it is easy to generate a large scale dataset of many different objects.

Specifically, we utilize the Tabletop Object Dataset generated from \cite{xie2019the} for training. This dataset consists of 40,000 synthetic scenes of cluttered objects on a tabletop in home environments. For each scene, a home environment is sampled from the SUNCG house dataset \cite{song2017semantic}, and a table and arbitrary objects are sampled from the ShapeNet dataset \cite{shapenet2015}. The number of objects for each scene is between 5 to 25. The PyBullet \cite{coumans2016pybullet} physics simulator is used to place objects on the table until they come to rest. After that, 7 RGB-D images are captured for each scene from different viewpoints. Fig.~\ref{fig:tabletop} shows some example images from the dataset.

\subsection{Learning RGB-D Feature Embeddings}

\begin{figure*}
	\centering
	\includegraphics[height=0.25\textwidth,width=\textwidth]{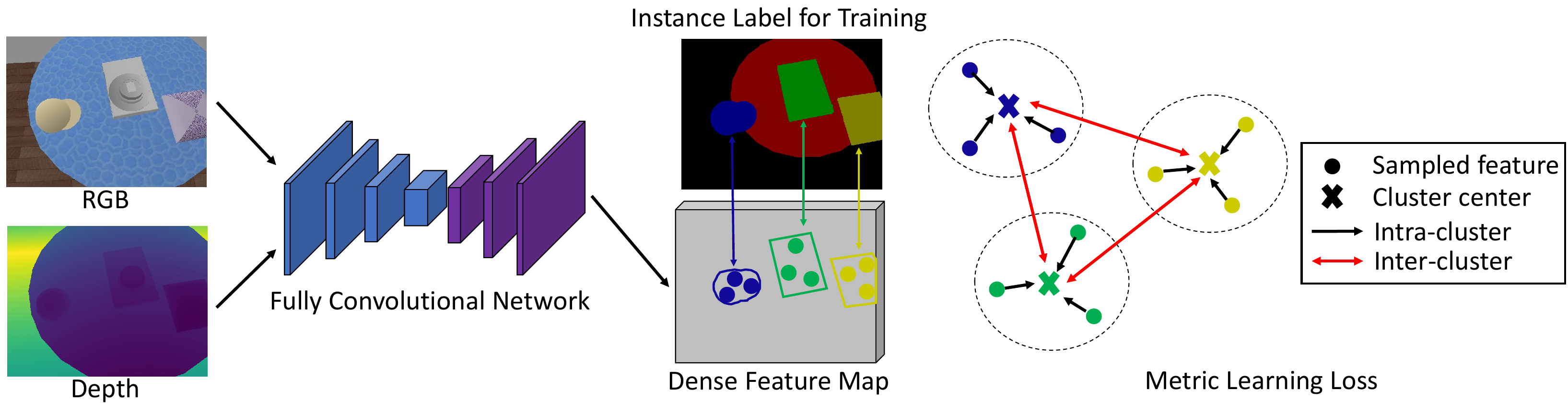}
	\caption{Illustration of our method for learning RGB-D feature embeddings using a fully convolutional network and a metric learning loss, where the loss function pushes pixels on the same object close to the cluster center and pushes the cluster centers of different objects far from each other in the embedding space.}
	\label{fig:network}
	\vspace{-6mm}
\end{figure*}
 
We can see that from Fig.~\ref{fig:tabletop}, the synthetic RGB images are non-photorealistic, which causes a domain gap when applying the trained network to real images. As a result, previous works \cite{xie2019the,danielczuk2019segmenting} mainly rely on using the depth images for segmentation, since it has been shown that networks trained on synthetic depth images can generalize to real images well \cite{mahler2017dex}. In our work, we investigate how to utilize these non-photorealistic synthetic RGB images combined with depth images to improve unseen object instance segmentation. Our solution is to learn RGB-D feature embeddings for clustering in a metric learning framework.

Specifically, given an RGB image $I \in \mathbb{R}^{H \times W \times 3}$ and depth image $D \in \mathbb{R}^{H \times W}$, where $H$ and $W$ are the image height and width, respectively, we first back-project the depth image $D$ into an ``organized'' point cloud $P \in \mathbb{R}^{H \times W \times 3}$ using the camera intrinsics. Then a Fully Convolutional Network (FCN) $\Psi$ takes the RGB image and the point cloud image as input, and computes a dense feature map $F = \Psi(I, P)  \in \mathbb{R}^{H \times W \times C}$, where $C$ is the dimension of the feature embeddings. Our FCN consists a backbone network and a set of deconvolutional layers to generate a dense feature map. Different backbone networks can be used such as VGG~\cite{simonyan2014very}, U-Net~\cite{ronneberger2015u} or ResNet~\cite{he2016deep}.

During inference, we simply apply a clustering algorithm after computing the feature map $F$ to group pixels together using their feature embeddings. Consequently, the goal of training is to make sure pixels from the same object are close to each other while pixels from different objects are far from each other in the embedding space. We apply a metric learning loss function to achieve this goal similar to \cite{de2017semantic,xie2019object}. First, suppose there are $K$ objects in the input image. For each object, we randomly sample $N$ pixels to compute the loss, where $N=1000$ in our experiments. Note that background is treated as one of the objects. We found that sampling the same number of pixels per object improves performance by balancing the importance of each object regardless of their sizes in the image. Second, assuming all feature embeddings are normalized to have unit length, we compute the spherical mean as defined in~\cite{xie2019object} for the $k^{\textrm{th}}$ object as
$\mu^k = \frac{\sum_{i=1}^N \mathbf{x}_i^k} {\| \sum_{i=1}^N \mathbf{x}_i^k \|}$, where $\mathbf{x}_i^k \in \mathbb{R}^C$ denotes the feature embedding of the $i^{\textrm{th}}$ pixel of the $k^{\textrm{th}}$ object. Third, we define the intra-cluster loss function as
\begin{equation} \label{eq:intra}
    \ell_{\textrm{intra}} = \frac{1}{K} \sum_{k=1}^K \sum_{i=1}^{N} \frac{\mathbbm{1}\left\{d(\mu^k, \mathbf{x}_i^k) - \alpha \geq 0\right\}\ d^2(\mu^k, \mathbf{x}_i^k)}{\sum_{i=1}^{N} \mathbbm{1}\left\{d(\mu^k, \mathbf{x}_i^k) - \alpha \geq 0\right\}},
\end{equation}
where $d(\mu^k, \mathbf{x}_i^k) = \frac{1}{2}(1 - \mu^k \cdot \mathbf{x}_i^k)$ is the cosine distance between the two unit-length feature vectors, $\alpha$ is the margin for the intra-cluster distance, and $\mathbbm{1}$ denotes the indicator function. The above intra-cluster loss function pushes feature embeddings of pixels on the same object close to the cluster center. Fourth, we define the inter-cluster loss function as
\begin{equation} \label{eq:inter}
    \ell_{\textrm{inter}} = \frac{2}{K(K-1)} \sum_{k < k'} \left[\delta - d(\mu^k, \mu^{k'}) \right]_+^2 \; ,
\end{equation}
where $[x]_+ = \max(x,0)$, and $\delta$ is the margin for the inter-cluster distance. The inter-cluster loss function pushes the cluster centers of different objects far away from each other in the embedding space. Finally, our loss function for learning RGB-D feature embeddings is $\mathcal{L} = \lambda_{\textrm{intra}} \ell_{\textrm{intra}} + \lambda_{\textrm{inter}} \ell_{\textrm{inter}}$, where we simply set $\lambda_{\textrm{intra}} = \lambda_{\textrm{inter}} = 1$ in our experiments. Fig.~\ref{fig:network} illustrates our network and loss function for learning RGB-D feature embeddings.

\subsection{Fusing RGB and Depth}

\begin{figure*}
	\centering
	\includegraphics[height=0.22\textwidth,width=\textwidth]{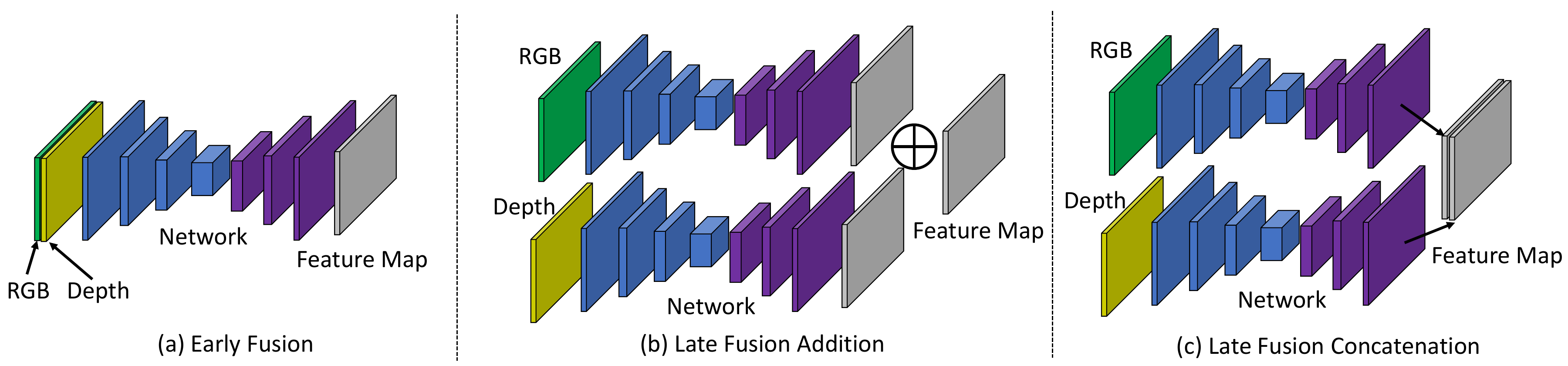}
	\caption{Three different ways of fusing RGB and depth in computing the feature embeddings.}
	\label{fig:fusion}
	\vspace{-4mm}
\end{figure*}

Fig.~\ref{fig:fusion} illustrates three different ways of fusing RGB and Depth (the point cloud image in our case) to compute the feature map. We will experiment with them for comparison. In Early Fusion, the RGB image $I$ and the point cloud image $P$ are concatenated before feeding them into the network. In this case, the input of the network is $\phi_{\text{early}}(I, P) \in \mathbb{R}^{H \times W \times 6}$. In Late Fusion Addition, $I$ and $P$ are fed into two towers of the same network architecture to compute two feature maps, and then the two feature maps are summed together. The two towers are trained to have different weights to account for the differences in color and depth. In Late Fusion Concatenation, the two feature maps are concatenated to generate a feature map $F \in \mathbb{R}^{H \times W \times 2C}$ instead.

\subsection{Mean Shift Clustering in Embedding Space}

We employ the mean shift algorithm~\cite{comaniciu2002mean} to cluster pixels in the embedding space due to its ability to automatically discover the number of clusters by seeking local maxima of the underlying probability
distribution. Furthermore, since our feature embeddings are normalized to have unit length, we apply the von Mises-Fisher mean shift (vMF-MS) algorithm~\cite{kobayashi2010mises}, where the von Mises-Fisher (vMF) distribution is used. Its probability density function is defined as
\begin{equation}
    p(\mathbf{x}; \mathbf{\mu}, \kappa) = C(\kappa) \exp ( \kappa \mathbf{x}^T \mu ),
\end{equation}
where both $\mathbf{x}$ and $\mu$ are unit-length vectors, $\mu$ indicates the center of the distribution, $\kappa$  controls the concentration of the distribution to the vector $\mu$, and $C(\kappa)$ is a normalization constant. Algorithm 1 summarizes the vMF-MS clustering algorithm. We reshape the feature map $F \in \mathbb{R}^{H \times W \times C}$ into a feature embedding matrix $\mathbf{X} \in \mathbbm{R}^{n \times C}$, where $n = H \times W$, and feed $\mathbf{X}$ into Algorithm 1.

\begin{algorithm}
\SetAlgoLined
\KwIn{ Feature embedding matrix $\mathbf{X} \in \mathbbm{R}^{n \times C}$, $\kappa$, $\epsilon$, number of seed $m$, number of iteration $T$}
 Sample $m$ initial clustering centers from $\mathbf{X}$ as the $m$ furthest points, denote it as $\mu^{(0)} \in \mathbbm{R}^{m \times C}$ \;
 \For{$t\leftarrow 1$ \KwTo $T$}{
  Compute weight matrix $\mathbf{W} \leftarrow \exp (\kappa \mu^{(t-1)} \mathbf{X}^T)$ \;
  
  Update $\mu^{(t)'} \leftarrow \mathbf{W} \mathbf{X}$\;
  Normalize each row vector in $\mu^{(t)'}$ to obtain $\mu^{(t)}$ \;
 }
 Merge cluster centers in $\mu^{(T)}$ with cosine distance smaller than $\epsilon$ \;
 Assign each pixel to the closest cluster center \;
 \caption{von Mises-Fisher mean shift clustering}
\end{algorithm}

\subsection{Zoom-in Cluster Refinement}

\begin{figure*}
	\centering
	\includegraphics[height=0.3\textwidth,width=0.95\textwidth]{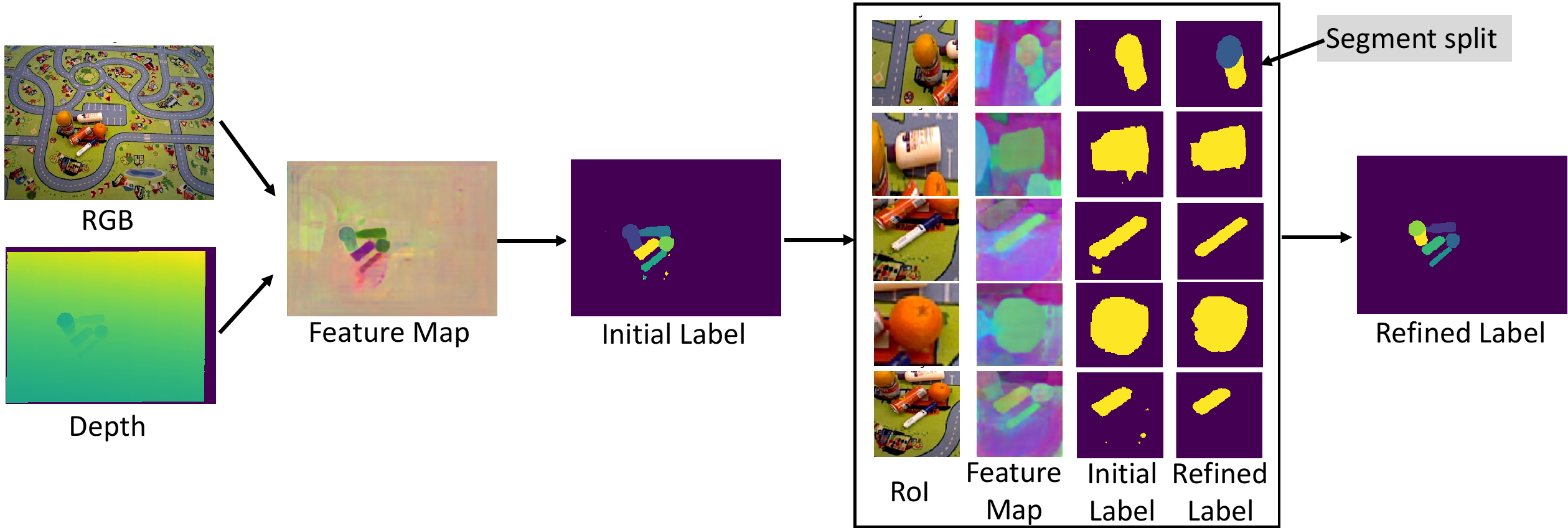}
	\caption{The two-stage clustering process in our method. The first stage clusters feature embeddings of all the image pixels. The second stage refines the segment for each RoI by clustering feature embeddings of the RoI.}
	\label{fig:refine}
	\vspace{-6mm}
\end{figure*}

We propose a two-stage clustering process to further improve the segmentation accuracy. The motivation is to tackle challenging scenes with objects close to each other or on top of each other. In these cases, clustering feature embeddings of all the image pixels may under-segment objects, since the network may treat two objects close to each other as one object. Therefore, we introduce a second stage of clustering, which we denote as \textit{zoom-in cluster refinement} as illustrated in Fig.~\ref{fig:refine}.

Specifically, for each cluster from the first stage, we crop an RGB-D Region of Interest (RoI), which is obtained by padding the cluster segment and resizing it to an image patch with size $224 \times 224$. Then, feature embeddings are computed for the RoI with an additional network which is trained with synthetic RoIs from the Tabletop Object Dataset. According to our experiments, training a separate network at the RoI-level is necessary. Directly applying the network trained on the whole images does not work well for the RoIs. Finally, the vMF-MS algorithm is used to cluster the feature embeddings of the RoIs. 

Since there could be multiple objects in an RoI, for each RoI, we only keep segments which have overlap larger than a predefined threshold (0.5 in our experiments) with the original segment from the first stage, i.e., the target segment to be refined. In this way, the target segment can be refined to have sharper boundary or divided into two separate objects in the under-segmented cases as shown in Fig.~\ref{fig:refine}. The final segmentation label for the whole image is computed by simply aggregating the segments from all the RoIs, and assigning a different object ID for each segment.

\vspace{-2mm}
\section{Experiments}
\vspace{-2mm}
\label{sec:experiment}

We first conduct ablation studies of several components in our method: Unseen Clustering Network (UCN). For comparison, we adopt Mask R-CNN \cite{he2017mask} which is trained to detect foreground objects with bounding boxes and then segment an object inside each bounding box. \cite{danielczuk2019segmenting} has successfully shown that Mask R-CNN achieves strong performance on segmenting unknown objects by training with synthetic depth images. After the ablation studies, we present the comparison of our method with the state-of-the-art methods for UOIS, and disucss failure cases from our method.

\subsection{Implementation Details}

Our network consists of a 34-layer, stride-8 ResNet (ResNet34-8s) with bilinearly upsampling to produce a full resolution 640x480 feature map with embedding dimension $C=64$. ResNet34-8s achieves better performance than networks using VGG or U-Nets as backbones in our experiments. Pretrained weights of ResNet34-8s on ImageNet \cite{deng2009imagenet} are used to initialize the RGB tower in Late Fusion Addition and Late Fusion Concatenation (Fig.~\ref{fig:fusion}(b)(c)). For early fusion and the depth tower in late fusion, no pretrained weights are used. The network is trained with the Adam optimizer \cite{kingma2014adam} for 16 epochs on the Tabletop Object Dataset with batch size 16 and learning rate 1e-5.

We train Mask R-CNN with the same number of epochs on the Tabletop Object Dataset using the SGD optimizer. We modified Mask R-CNN to support training with RGB-D images, where the two-tower structure is used in late fusion. Similarly, pretrained weights on ImageNet are only used for the RGB tower in Mask R-CNN. 

In our metric learning loss, we set margin $\alpha=0.02$ in the intra-cluster loss, and margin $\delta=0.5$ in the inter-cluster loss. In the von Mises-Fisher mean shift clustering, we set $\kappa=20$ and $\epsilon=2\alpha$. We run the clustering for 10 iterations on 100 initial seeds. The run time of our method is on average 0.2s for the 1st stage and 0.05s per object in the 2nd stage on a TITAN Xp GPU for 640x480 images.

\begin{table*}[t]
\resizebox{\linewidth}{!}{\begin{tabular}{|c|c|ccc|ccc|c||ccc|ccc|c|}
\hline
\multirow{3}{*}{Method} & \multirow{3}{*}{Input} & \multicolumn{7}{c||}{OCID \cite{suchi2019easylabel} (2390 images)}  & \multicolumn{7}{c|}{OSD \cite{richtsfeld2012segmentation} (111 images) } \\ \cline{3-16}
 &  & \multicolumn{3}{c|}{Overlap} & \multicolumn{3}{c|}{Boundary} & & \multicolumn{3}{c|}{Overlap} & \multicolumn{3}{c|}{Boundary} & \\
 &  & P & R & F & P & R & F & $\%75$ & P & R & F & P & R & F & $\%75$ \\ \hline
MRCNN & RGB & 77.6 & 67.0 & 67.2 & 65.5 & 53.9 & 54.6 & 55.8 & 64.2 & 61.3 & 62.5 & 50.2 & 40.2 & 44.0 & 31.9\\
UCN (Ours) & RGB & 54.8 & 76.0 & 59.4 & 34.5 & 45.0 & 36.5 & 48.0 & 57.2 & 73.8 & 63.3 & 34.7 & 50.0  & 39.1 & 52.5 \\
\hline
MRCNN & Depth & 85.3 & 85.6 & 84.7 & \textbf{83.2} & 76.6 & \textbf{78.8} & 72.7 & 77.8 & 85.1 & 80.6 & 52.5 & 57.9  & 54.6 & 77.6\\
UCN (Ours) & Depth & 83.1 & 90.7 & 86.4 & 77.7 & 74.3 & 75.6 & 75.4 & 78.7 & 83.8 & 81.0 & 52.6 & 50.0  & 50.9 & 72.1 \\
\hline
MRCNN & RGBD early & 78.7 & 79.0 & 78.1 & 73.4 & 70.3 & 70.8 & 62.2 & 78.3 & 78.4 & 78.3 & 65.2 & 62.2  & 63.2 & 61.2 \\
UCN (Ours) & RGBD early & 78.8 & 89.2 & 82.8 & 66.9 & 69.7 & 67.2 & 73.5 & 77.4 & 81.8 & 79.2 & 53.9 & 53.0  & 53.0 & 69.0 \\
\hline
MRCNN & RGBD add & 79.6 & 76.7 & 76.6 & 68.7 & 63.7 & 64.3 & 62.9 & 66.4 & 64.8 & 65.5 & 53.7 & 43.8  & 47.5 & 37.1 \\
UCN (Ours) & RGBD add & \textbf{86.0} & \textbf{92.3} & \textbf{88.5} & 80.4 & \textbf{78.3} & \textbf{78.8} & \textbf{82.2} & \textbf{84.3} & \textbf{88.3} & \textbf{86.2} & \textbf{67.5} & \textbf{67.5}  & \textbf{67.1} & \textbf{79.3} \\
\hline
MRCNN & RGBD concat & 79.6 & 76.2 & 76.0 & 68.2 & 63.5 & 63.7 & 63.0 & 67.0 & 63.8 & 65.3 & 53.1 & 42.7 & 46.5 & 37.1 \\
UCN (Ours) & RGBD concat & 79.2 & 87.8 & 82.9 & 70.6 & 67.5 & 68.5 & 68.3 & 76.4 & 83.3  & 79.7 & 50.5 & 48.5  & 48.8 & 67.5 \\
\hline

\end{tabular}}
\caption{Evaluation of our method and Mask R-CNN~\cite{he2017mask} trained on different input modes.}
\label{table:fusion}
\vspace{-6mm}
\end{table*}

\vspace{-1mm}
\subsection{Datasets and Evaluation Metrics}
\vspace{-1mm}

We evaluate our methods on the Object Clutter Indoor Dataset (OCID) \cite{suchi2019easylabel} and the Object Segmentation Database (OSD) \cite{richtsfeld2012segmentation} for UOIS in tabletop scenes. OCID contains 2,390 RGB-D images with up to 20 objects in an image, while OSD has 111 RGB-D images with up to 15 objects in an image. On average, there are 7.5 objects per image in OCID and 3.3 objects per image in OSD. Both datasets contain challenging images of cluttered scenes. The main difference between the two datasets is that, the images in OSD are manually annotated, while the annotations in OCID are obtained in a semi-automatic way by adding objects to each scene one by one and then checking the depth differences. As a result, the segmentation masks from OCID are prone to the noise from the depth images. We use all images from the two datasets for testing.

Following \cite{xie2019the}, we use precision, recall and F-measure to evaluate the object segmentation performance. These three metrics are first computed between all pairs of predicted objects and ground truth objects. Then the Hungarian method with pairwise F-measure is used to compute a matching between predicted objects and ground truth. Given this matching, the final precision, recall and F-measure are computed by $P = \frac{\sum_i \left|c_i \cap g(c_i) \right|}{\sum_i \left|c_i\right|}$, $R = \frac{\sum_i \left|c_i \cap g(c_i) \right|}{\sum_j \left|g_j\right|}$, $F = \frac{2PR}{P+R}$, where $c_i$ denotes the set of pixels belonging to predicted object $i$, $g(c_i)$ is the set of pixels of the matched ground truth object of $c_i$, and $g_j$ is the set of pixels for ground truth object $j$.

We denote the above three metrics as Overlap P/R/F since the true positives are counted by the pixel overlap of the whole object. \cite{xie2019the} also introduces the Boundary P/R/F to evaluate how sharp the predicted boundary matches against the ground truth boundary, where the true positives are counted by pixel overlap of the two boundaries. Additionally, we employ the percentage of segmented objects with Overlap F-measure $\geq 75\%$ as a metric to indicate the percentage of objects that can be
segmented with a certain accuracy~\cite{ochs2013segmentation}.

\vspace{-1mm}
\subsection{Effect of Input Mode}
\vspace{-1mm}

We first study the effect of different input modes in training our method and Mask R-CNN on the Tabletop Object Dataset for UOIS. We train different networks of both methods using RGB, depth, and RGB-D images as input. When using RGB-D images, three different ways for fusing RGB and depth as shown in Fig.~\ref{fig:fusion} are investigated. 

\begin{figure*}
	\centering
	\includegraphics[height=0.21\textwidth,width=0.95\textwidth]{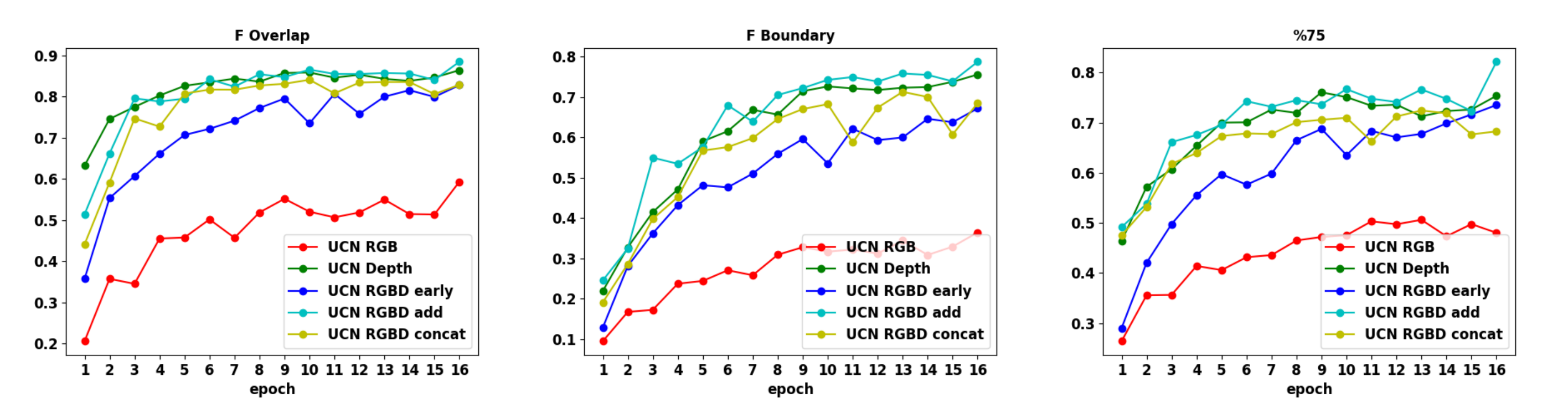}
	\caption{Learning curves of our method on the OCID dataset in terms of different number of training epochs.}
	\label{fig:epochs}
	\vspace{-2mm}
\end{figure*}

Table~\ref{table:fusion} presents the segmentation results on OCID and OSD. From the table, we can see that i) using depth information significantly boosts the performance for both methods. This is mainly because training with the non-photorealistic RGB images only does not work well for real world images; ii) For Mask R-CNN, using depth as input achieves the best performance. Using RGB-D images as input is worse than just using depth for any of the three fusion ways; iii) For our method, using RGB-D images with Late Fusion Addition achieves the best performance, which is also better than Mask R-CNN trained on depth. This result demonstrates that, unlike Mask R-CNN, our method is able to utilize the non-photorealistic RGB images in addition to the depth images to learn powerful feature embedding networks that can be effectively transferred to real world images. Fig.~\ref{fig:epochs} shows the learning curves of our method in different input modes on the OCID dataset.

\vspace{-2mm}
\subsection{Effect of the Two-stage Clustering Algorithm}
\vspace{-1mm}

\begin{table*}[t]
\resizebox{\linewidth}{!}{\begin{tabular}{|c|ccc|ccc|c||ccc|ccc|c|}
\hline
\multirow{3}{*}{UCN (Ours)} & \multicolumn{7}{c||}{OCID \cite{suchi2019easylabel} (2390 images)}  & \multicolumn{7}{c|}{OSD \cite{richtsfeld2012segmentation} (111 images) } \\ \cline{2-15}
 &  \multicolumn{3}{c|}{Overlap} & \multicolumn{3}{c|}{Boundary} & & \multicolumn{3}{c|}{Overlap} & \multicolumn{3}{c|}{Boundary} & \\
 & P & R & F & P & R & F & $\%75$ & P & R & F & P & R & F & $\%75$ \\ \hline
RGB & 54.8 & 76.0 & 59.4 & 34.5 & 45.0 & 36.5 & 48.0 & 57.2 & 73.8 & 63.3 & 34.7 & 50.0  & 39.1 & 52.5 \\
RGB + Zoom-in & 59.1 & 74.0 & 61.1 & 40.8 & 55.0 & 43.8 & 58.2 & 59.1 & 71.7  & 63.8 & 34.3 & 53.3 & 39.5 & 52.6 \\
\hline
Depth & 83.1 & 90.7 & 86.4 & 77.7 & 74.3 & 75.6 & 75.4 & 78.7 & 83.8 & 81.0 & 52.6 & 50.0  & 50.9 & 72.1 \\
Depth + Zoom-in & 88.7 & 92.2 & 90.1 & 83.6 & 84.2 & 83.3 & 85.3 & 83.5 & 86.0  & 84.5 & 57.2 & 56.8 & 56.5 & 80.6  \\
\hline
RGBD early & 78.8 & 89.2 & 82.8 & 66.9 & 69.7 & 67.2 & 73.5 & 77.4 & 81.8 & 79.2 & 53.9 & 53.0  & 53.0 & 69.0 \\
RGBD early + Zoom-in & 85.2 & 90.3 & 86.8 & 73.5 & 79.9 & 75.3 & 84.3 & 79.4 & 79.7  & 79.2 & 57.3 & 58.6 & 57.1 & 67.9 \\
\hline
RGBD add & 86.0 & 92.3 & 88.5 & 80.4 & 78.3 & 78.8 & 82.2 & 84.3 & \textbf{88.3} & 86.2 & 67.5 & 67.5  & 67.1 & 79.3 \\
RGBD add + Zoom-in & \textbf{91.6} & \textbf{92.5} & \textbf{91.6} & \textbf{86.5} & \textbf{87.1} & \textbf{86.1} & \textbf{89.3} & \textbf{87.4} & 87.4 & \textbf{87.4} & \textbf{69.1} & \textbf{70.8} & \textbf{69.4} & \textbf{83.2} \\
\hline
RGBD concat & 79.2 & 87.8 & 82.9 & 70.6 & 67.5 & 68.5 & 68.3 & 76.4 & 83.3  & 79.7 & 50.5 & 48.5  & 48.8 & 67.5 \\
RGBD concat + Zoom-in & 88.1 & 90.1 & 88.8 & 82.1 & 83.1 & 82.0 & 82.4 & 82.0 & 85.0  & 83.4 & 57.8 & 59.2 & 57.9 & 76.5 \\
\hline
\end{tabular}}
\caption{Evaluation of the proposed zoom-in cluster refinement algorithm on different input modes.}
\label{table:refine}
\vspace{-7mm}
\end{table*}

\begin{figure}
	\centering
	\includegraphics[height=0.44\textwidth,width=0.95\textwidth]{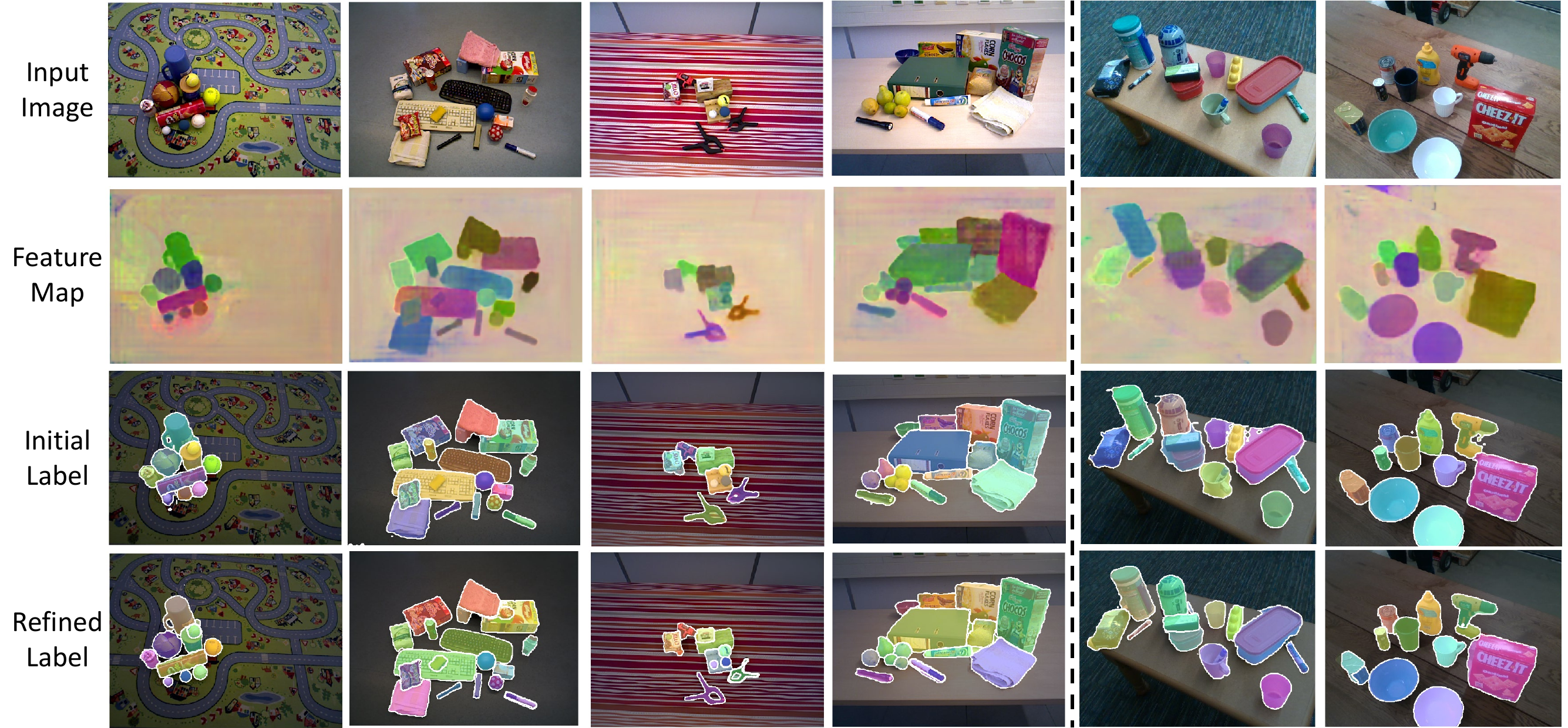}
	\caption{Examples of the feature maps, initial labels from the first stage clustering and the refined labels after the second stage clustering from our method. The last two columns show results on images from our lab.}
	\label{fig:result_refine}
	\vspace{-4mm}
\end{figure}

We investigate the effect of our two-stage clustering algorithm to see how much the zoom-in cluster refinement can help. Table~\ref{table:refine} shows the results before and after refinement for the five different input modes. As we can see from the table, the zoom-in cluster refinement algorithm significantly improves the Boundary P/R/F and the percentage of objects with F-measure larger than 0.75 on both datasets. This result confirms that the second stage clustering generates sharper object boundaries and helps separate objects that are under-segmented from the first stage. For the Overlap P/R/F, the refinement improves in most cases, but scarifies a small amount of recall in the trade off of better precision for some models.

Fig.~\ref{fig:result_refine} shows some qualitative results of our method on OCID and images from our lab using RGB-D images as input with Late Fusion Addition. We can see how the object boundaries are refined and objects are correctly separated from each other after refinement in these examples. To visualize the feature map with dimension 64, we first sum along channels with interval three to obtain an image with dimension 3 and then normalize all the values between 0 and 255 for visualization. We can clearly see that from the feature map, objects are separated in the embedding space.

\subsection{Comparison to the State-of-the-art Methods}

\begin{table*}[t]
\resizebox{\linewidth}{!}{\begin{tabular}{|c|ccc|ccc|c||ccc|ccc|c|}
\hline
\multirow{3}{*}{Method} & \multicolumn{7}{c||}{OCID \cite{suchi2019easylabel} (2390 images)}  & \multicolumn{7}{c|}{OSD \cite{richtsfeld2012segmentation} (111 images) } \\ \cline{2-15}
 &  \multicolumn{3}{c|}{Overlap} & \multicolumn{3}{c|}{Boundary} & & \multicolumn{3}{c|}{Overlap} & \multicolumn{3}{c|}{Boundary} & \\
 &  P & R & F & P & R & F & $\%75$ & P & R & F & P & R & F & $\%75$ \\ \hline

\hline
MRCNN Depth~\cite{he2017mask} & 85.3 & 85.6 & 84.7 & 83.2 & 76.6 & 78.8 & 72.7 & 77.8 & 85.1 & 80.6 & 52.5 & 57.9  & 54.6 & 77.6\\

UOIS-Net-2D~\cite{xie2019the} & 88.3 & 78.9 & 81.7 & 82.0 & 65.9 & 71.4 & 69.1 & 80.7 & 80.5 & 79.9 & 66.0 & 67.1 & 65.6 & 71.9 \\

UOIS-Net-3D~\cite{xie2020unseen} & 86.5 & 86.6 & 86.4 & 80.0 & 73.4 & 76.2 & 77.2 & 85.7 & 82.5 & 83.3 & \textbf{75.7} & 68.9 & \textbf{71.2} & 73.8 \\

UCN (Ours) & \textbf{91.6} & \textbf{92.5} & \textbf{91.6} & \textbf{86.5} & \textbf{87.1} & \textbf{86.1} & \textbf{89.3} & \textbf{87.4} & \textbf{87.4} & \textbf{87.4} & 69.1 & \textbf{70.8} & 69.4 & \textbf{83.2} \\

\hline

\end{tabular}}
\caption{Comparison with the state-of-the-art methods for unseen object instance segmentation.}
\label{table:comparison}
\vspace{-6mm}
\end{table*}

We compare our method to the state-of-the-art methods for UOIS in the literature. We use our best model: the network trained with RGB-D images in Late Fusion Addition plus the zoom-in refinement. Table~\ref{table:comparison} presents the results. Our approach achieves the best Overlap F-measure and Boundary F-measure on the OCID dataset, and the best Overlap F-measure on the OSD dataset. Additionally, we significantly outperform the other methods on the percentage of objects segmented with F-measure larger than 0.75. Note that all methods are trained on the Tabletop Object Dataset.

The main difference between our method and Mask R-CNN \cite{he2017mask} and UOIS-Net \cite{xie2020unseen} is that, our method is a bottom-up approach and we learn a distance metric in the embedding space to compare pixels. Both Mask R-CNN and UOIS-Net have softmax functions to classify pixels into foreground and background. As a result, our method can utilize the non-photorealistic synthetic RGB images to help learning the distance metric, while preventing overfitting to the synthetic images. Due to the nature of a bottom-up approach in grouping pixels together, our method obtains higher recall compared to Mask R-CNN and UOIS-Net. 

\subsection{Discussion of Failure Cases}

\begin{figure*}[h]
	\centering
	\includegraphics[height=0.25\textwidth,width=0.95\textwidth]{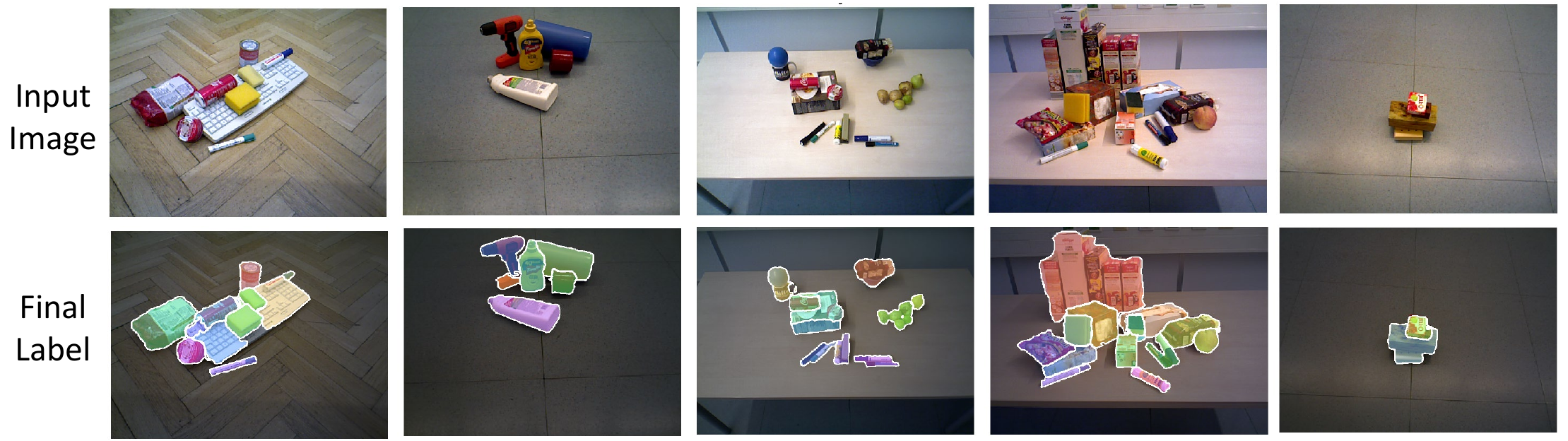}
	\caption{Examples of segmentation failure cases from our method.}
	\label{fig:result_failure}
	\vspace{-2mm}
\end{figure*}

We discuss segmentation failure cases from our method and show some examples in Fig.~\ref{fig:result_failure}. The majority failures are under-segmentation. In some cases, our method cannot separate similar objects that are very close to each other or on top of each other, even with the zoom-in refinement step. As we can see in Fig.~\ref{fig:result_failure}, the method fails to separate two pens, a group of similar fruits or a row of boxes in these examples. Another failure is over-segmentation, and we see this mainly happens for the power drill and some keyboards in the OCID dataset. Future work could investigate how to fix there failures using multi-view images or active perception with robots.


\section{Conclusion}
\label{sec:conclusion}

We have introduced a simple but effective approach for UOIS by learning RGB-D feature embeddings from synthetic data. Different from previous approaches that mainly rely on synthetic depth for segmentation, our approach can utilize the non-photorealistic RGB images jointly with the depth images to learn a distance metric for clustering. We also introduce a new two-stage clustering algorithm to tackle challenging scenes when objects are close to each other. Our experiments demonstrate the effectiveness of our approach for UOIS on real-world images.

\clearpage


\bibliography{example}  

\end{document}